\documentclass{article}
\usepackage{amssymb}
\usepackage{amsmath}
\usepackage{booktabs}
\usepackage{graphicx}
\usepackage{hyperref}
\usepackage{natbib}

\makeatletter
\renewcommand\hyper@natlinkbreak[2]{#1}
\makeatother

\usepackage{iclr_fmwild}
\iclrfinalcopy
\usepackage{times}
\usepackage{filecontents}

\begin{document}
\title{Compositional Subspace Representation Fine-tuning for Adaptive Large Language Models}
\author{Andy Zhou \\
Intology AI
}
\maketitle

\begin{abstract}
Adapting large language models to multiple tasks can cause cross-skill interference, where improvements for one skill degrade another. While methods such as LoRA impose orthogonality constraints at the weight level, they do not fully address interference in hidden-state representations. We propose Compositional Subspace Representation Fine-tuning (CS-ReFT), a novel representation-based approach that learns multiple orthonormal subspace transformations, each specializing in a distinct skill, and composes them via a lightweight router. By isolating these subspace edits in the hidden state, rather than weight matrices, CS-ReFT prevents cross-task conflicts more effectively. On the AlpacaEval benchmark, applying CS-ReFT to Llama-2-7B achieves a 93.94\% win rate, surpassing GPT-3.5 Turbo (86.30\%) while requiring only 0.0098\% of model parameters. These findings show that specialized representation edits, composed via a simple router, significantly enhance multi-task instruction following with minimal overhead.
\end{abstract}

\section{Introduction}

Large language models (LLMs) have become central to a wide range of NLP applications, yet adapting them to new tasks can be computationally expensive, often requiring hundreds of GPU hours and significant memory overhead. Parameter-efficient fine-tuning (PEFT) methods \citep{Han2024parfin} tackle this challenge by updating only a small fraction of model parameters, typically 0.1--1\% of the total. While this approach has enabled more practical deployment of adapted models, with methods like LoRA \citep{Hu2021LoRALA} reducing parameter counts by 1000x, current PEFT techniques still focus primarily on \emph{weight-based} updates. In contrast, \emph{representation editing} methods like ReFT \citep{Wu2024refrep} directly modify hidden states, achieving even lower parameter overhead; however, most have used a single global edit that struggles to handle multiple skills without interference.

A core problem in multi-task adaptation is \emph{cross-task interference}, wherein changes aimed at improving one task degrade performance on another \citep{Pfeiffer2023moddee}. Although recent LoRA variants impose orthogonality constraints to reduce conflicts \citep{Wang2023OrthogonalSL,Hsu2024saflor}, none have extended these ideas to \emph{representation-based} fine-tuning, where orthonormal subspaces can isolate skills more effectively at the hidden-state level. To address this gap, we propose \textbf{Compositional Subspace Representation Fine-tuning (CS-ReFT)}, a framework that extends ReFT with multiple orthonormal subspace edits and a lightweight router for dynamic composition. Our contributions include:
\begin{itemize}
    \item We learn separate low-rank transformations for each skill implicitly, preventing conflicts across tasks while requiring only 0.0098\% of model parameters--a 12.7x reduction compared to LoRA. A small gating network is trained to selectively activate relevant subspaces for each input.
    \item By applying orthonormal constraints \emph{directly in hidden-state space}, CS-ReFT isolates skills more cleanly than weight-based orthogonal LoRA methods.
    \item CS-ReFT attains a 93.94\% win rate on AlpacaEval with Llama-2-7B--significantly outperforming both larger models (GPT-3.5-Turbo) and parameter-efficient baselines (LoReft at 85.60\% ).
\end{itemize}

\begin{figure}[t]
    \centering
    \includegraphics[width=\linewidth]{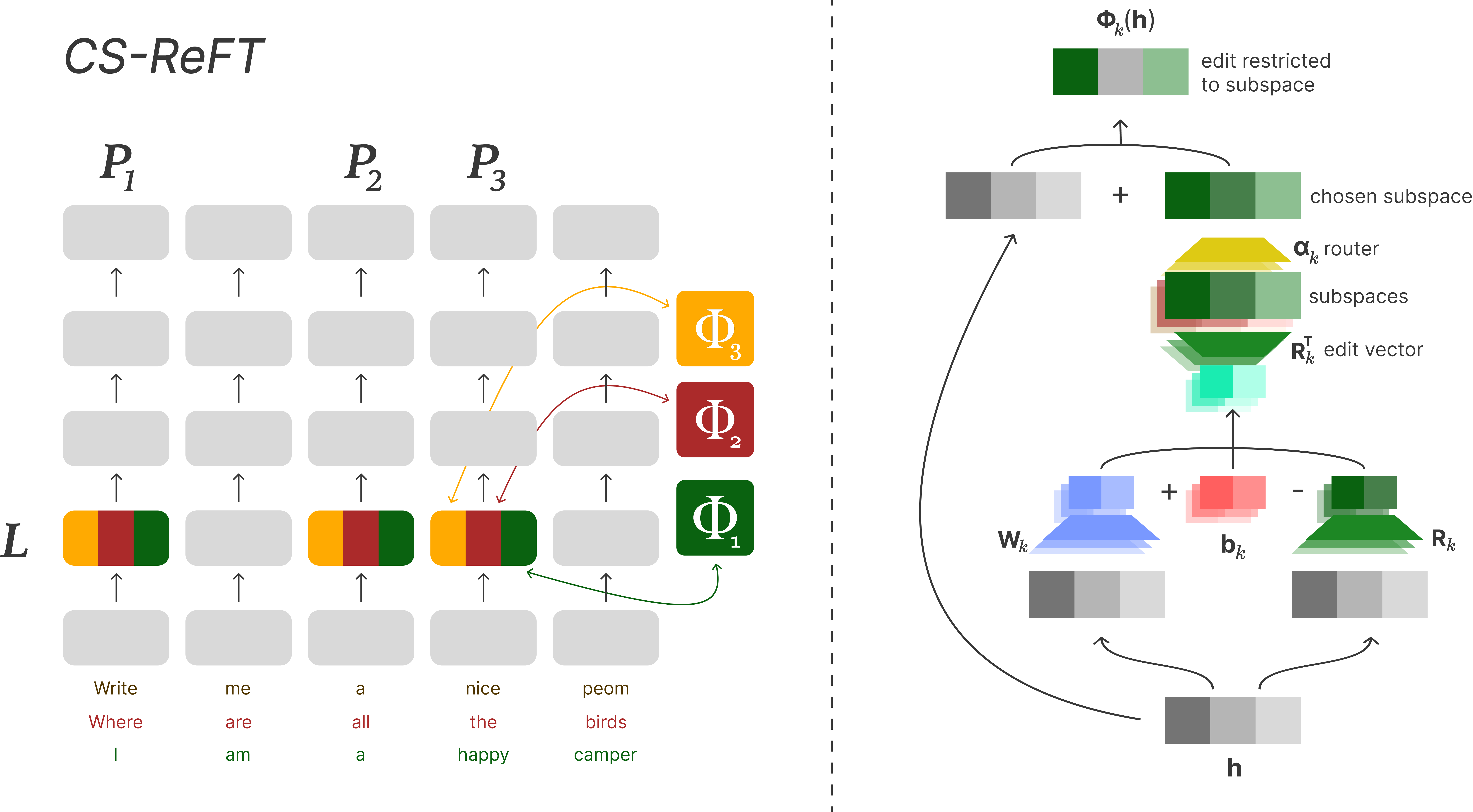}
    \caption{\textbf{Illustration of CS-ReFT.} (1) The left panel shows how Compositional Subspace Representation Fine-Tuning (CS-ReFT) applies specialized subspace transformations (\(\Phi_1, \Phi_2, \Phi_3\)) at specific positions in different layers to adapt a frozen model for multiple tasks. Each subspace edit is task-specific, reducing interference while allowing composition when needed. (2) The right panel details the routing mechanism: a lightweight router determines which subspaces to activate based on the input, ensuring efficient and targeted modifications.}
    \label{fig:enter-label}
\end{figure}

\section{Related Work}
\label{sec:related}

\paragraph{Parameter-efficient adaptation.} Recent years have seen rapid progress in parameter-efficient fine-tuning (PEFT) methods \citep{Han2024parfin,Lialin2023scadow,Li2021PrefixTuningOC}. Low-rank adaptation approaches like LoRA \citep{Hu2021LoRALA} decompose weight updates into low-rank matrices \(U V^\top\), typically achieving 1000x parameter reduction while maintaining performance. Other methods like BitFit \citep{Ben-Zaken2021bitsim} modify only bias terms. Representation-based methods \citep{Wu2024refrep,Kong2024alilar,Zou2023RepresentationEA} instead edit model activations. These advances have made LLM adaptation more practical, achieving lower parameter overhead than weight-based updates, but use a \emph{single} global edit function, limiting their effectiveness for multi-task adaptation.

\paragraph{Multi-task learning.} Multi-task adaptation strategies span several approaches, from shared parameter methods \citep{Hu2021LoRALA,Liu2024adafin} that risk interference, to task-specific modules \citep{Chronopoulou2022lanada,Yang2024mtllow} requiring separate adapters, to dynamic routing systems \citep{Araujo2024leato,Zhang2024mileff} that often introduce significant overhead. Some recent work combines orthogonality constraints with multi-task setups \citep{Hsu2024saflor,Liu2023divthe,Wang2023OrthogonalSL}, but again these rely on weight-based modules inserted inside Transformer layers. By contrast, \emph{our method applies orthonormal constraints at the \emph{representation level}}, learning \emph{disjoint subspaces} in the hidden state and dynamically routing between them. This design reduces cross-task interference compared to sharing a single low-rank factorization for all tasks. 

\section{Method}
\label{sec:method}

\textbf{Compositional Subspace Representation Fine-tuning (CS-ReFT)} learns \emph{multiple} low-rank \emph{subspace interventions} and a \emph{router} to activate them on a per-input basis, addressing cross-task interference by dedicating \emph{separate} subspaces to each skill. We selectively compose these subspaces at inference. Let \(\mathcal{M}\) be a frozen pretrained model (e.g., a Transformer) of hidden dimension \(d\). For each sequence of \(n\) tokens \(x = (x_1,\dots,x_n)\), the model produces \(\{\,h_1^{(j)},\dots,h_n^{(j)}\}\) in layer \(j\). Our goal is to adapt \(\mathcal{M}\) to a set of \(k\) tasks \(\{\mathcal{T}_1,\dots,\mathcal{T}_k\}\) \emph{without} modifying the original weights. Instead, we learn: (1) A collection of \emph{low-rank subspace transformations}, \(\{\Phi_1, \dots, \Phi_k\}\), one per task, (2) a \emph{router} \(R\) that decides which subset of \(\{\Phi_i\}\) to activate given an input. Our design ensures that each task \(\mathcal{T}_i\) has a dedicated subspace edit \(\Phi_i\)---preventing direct interference---yet also enables composition for inputs requiring multiple skills. In practice, the tasks are manually defined through manually partitioning the data or implicitly learned during training.

\label{sec:subspace}

\subsection{Subspace Representation Editing}
Following ReFT~\citep{Wu2024refrep}, each subspace intervention \(\Phi\) modifies a hidden vector \(h \in \mathbb{R}^d\) by editing only an \(r\)-dimensional subspace spanned by the rows of \(R\). Concretely, we let
\[
\Phi(h) 
=\;
h 
\;+\;
R^\top\bigl(\underbrace{W\,h + b}_{\text{desired subspace coords}} \;-\; R\,h\bigr),
\]
where
\(\;R\in\mathbb{R}^{r\times d}\)\; is typically constrained to have orthonormal rows (\(R\,R^\top = I_r\)), and \(W\in \mathbb{R}^{r\times d},\,b\in\mathbb{R}^r\) are trainable parameters. In CS-ReFT, we have \(\Phi_1,\dots,\Phi_k\), \emph{one per task}, each with its own low-rank parameters \(\{R_i,W_i,b_i\}\). Ensuring that an input requiring task \(i\) can be edited by \(\Phi_i\) without altering another subspace, this fully separates the learned directions in hidden space, mitigating interference across tasks.

\subsection{Router Mechanism}
Not every input belongs to a single task, nor do we want to dedicate a distinct subspace for every fine-grained skill. Hence, we introduce a router that selects or composes the relevant subspaces at inference time. For example, an instruction might require both \(\Phi_2\) (arithmetic) and \(\Phi_3\) (sentiment analysis). We define a small routing network 
\[
\mathrm{Router}(x) = \alpha \in [0,1]^{\{k\}},
\]
which maps an embedding of the input (e.g., the first token’s hidden state) to a gating vector \(\alpha\). We then compose the subspace edits as:
\[
h' 
=\; h
\;+\;
\sum_{i=1}^k\, \alpha_i\;\Bigl[\;
    R_i^\top\bigl(W_i\,h + b_i - R_i\,h\bigr)
\Bigr].
\]
If \(\alpha_i\) is discrete (e.g.\ thresholded), then each \(\Phi_i\) is \emph{on} or \emph{off}.  Alternatively, we can keep \(\alpha_i\in [0,1]\) for a soft gating.  In either case, the parameter overhead from the router is minimal, allowing dynamic composition without losing efficiency. Crucially, this router is \emph{jointly trained} alongside the subspaces themselves. As a result, the model can \emph{implicitly} discover how to route different inputs to different subspaces without any manual task partitioning.

\subsection{Training Objective}

We train CS-ReFT by minimizing:
\[
\mathcal{L} 
=\; \sum_{i=1}^k
\;\mathbb{E}_{(x,y)\sim\mathcal{T}_i}
\Bigl[\,\ell\bigl(\mathcal{M}\bigl(x;\{\Phi_i\}, R\bigr),\,y\bigr)\Bigr]
\;+\;\lambda\,\Omega(\alpha),
\]
where \(\ell(\cdot)\) is a task loss (e.g.\ cross-entropy), and \(\Omega(\alpha)\) can be a sparsity regularizer on the router outputs to encourage minimal subspace usage.  In practice, we update only \(\{\Phi_1,\dots,\Phi_k\}\) and the router’s parameters while leaving all original model weights frozen. This design prevents cross-task interference by activating only relevant subspaces on each input, and the low-rank structure keeps parameter overhead minimal.

In addition to these aspects, CS-ReFT provides multiple benefits. It prevents cross-task interference by keeping each skill’s subspace disjoint so that changes to \(\Phi_i\) do not overwrite \(\Phi_j\). It also fosters compositional synergy, as the router composes subspaces on demand to enable multi-skill prompts. Finally, it ensures extreme parameter savings because each subspace \(\Phi_i\) remains low-rank and the router is tiny, resulting in significantly fewer parameters than typical multi-head adapters. This compositional subspace design thus unifies the \emph{efficiency} of representation editing with the \emph{modularity} of multi-task routing, enabling high-quality, multi-task LLM adaptation with minimal overhead.

\section{Experiments}

\textit{Setup.} We evaluate CS-ReFT using the AlpacaEval benchmark \citep{Dubois2024LengthControlledAA}, which measures instruction-following capabilities through win rates against reference responses. As a general task, instruction-following implicitly involves multiple subtasks, such as reasoning and common-sense understanding. Our experiments use Llama-2-7B \citep{Touvron2023Llama2O} as the base model, comparing CS-ReFT against both parameter-efficient methods and larger models. We evaluate using two metrics: \textbf{Win Rate} (percentage of model outputs preferred over reference responses) and \textbf{Parameter Efficiency} (percentage of trainable parameters relative to full model). Our baselines include parameter-efficient methods (LoRA \citep{Hu2021LoRALA}, RED \citep{Wu2024AdvancingPE}, DiReFT \citep{Wu2024refrep}, LoReFT \citep{Wu2024refrep}) and larger models (GPT-3.5 Turbo \citep{Brown2020LanguageMA}, Llama-2-13B \citep{Touvron2023Llama2O}).

The CS-ReFT architecture implements four distinct low-rank transformations using the ReFT intervention mechanism \citep{Wu2024refrep}, each operating independently on the model's hidden states. A lightweight two-layer router network processes the first token's embedding ($h \in \mathbb{R}^d$), with an input layer mapping $\mathbb{R}^d \rightarrow \mathbb{R}^{d/2}$ (ReLU activation) and an output layer mapping $\mathbb{R}^{d/2} \rightarrow \mathbb{R}^4$ (sigmoid activation), using a 0.5 threshold for binary gating.

\begin{table}[t]
\centering
\caption{Performance on AlpacaEval. Parameter Efficiency (PE) shows fraction of trainable parameters relative to the base model. Win rate measures preference over reference responses. CS-ReFT on Llama-2-7B outperforms all baseline methods and is competitive with ReFT on parameter efficiency.}
\label{tab:main_results}
\begin{tabular}{lrr}
\toprule
\textbf{Model} & \textbf{Win Rate (\%)} & \textbf{PE (\%)} \\
\midrule
\addlinespace[0.4em]
\multicolumn{3}{l}{\textit{Reference Models}} \\
\addlinespace[0.3em]
GPT-3.5 Turbo 1106 & 86.30 & --- \\
Llama-2 Chat 13B & 81.10 & --- \\
Llama-2 Chat 7B & 71.40 & --- \\
\midrule
\addlinespace[0.4em]
\multicolumn{3}{l}{\textit{Parameter-Efficient Methods (Llama-2 7B)}} \\
\addlinespace[0.3em]
Full Fine-tuning & 80.93 & 100.00 \\
LoRA & 81.48 & 0.1245 \\
RED & 81.69 & 0.0039 \\
DiReFT & 84.85 & 0.0039 \\
LoReFT & 85.60 & \textbf{0.0039} \\
\midrule
CS-ReFT (Ours) & \textbf{93.94} & 0.0098 \\
\bottomrule
\end{tabular}
\end{table}

\textit{Results.} Table~\ref{tab:main_results} presents performance comparisons across model sizes and adaptation methods. CS-ReFT achieves a 93.94\% win rate while modifying only 0.0098\% of model parameters. Specifically, it surpasses larger models such as GPT-3.5 Turbo (86.30\%) and Llama-2-13B (81.10\%), outperforms weight-based methods like LoRA (81.48\%, 0.1245\% parameters), and exceeds representation methods such as ReFT variants (81.69--85.60\%, 0.0039\% parameters), highlighting the effectiveness of specialized subspaces and dynamic routing.

\section{Conclusion}
We introduced Compositional Subspace Representation Fine-tuning (CS-ReFT), which addresses cross-task interference by assigning separate low-rank subspace transformations to each skill and using a lightweight router for dynamic composition. Unlike orthonormal LoRA variants that still operate on weight matrices, our approach enforces \emph{orthonormal subspace constraints} directly on hidden states, thereby isolating learned features more effectively. Experiments on AlpacaEval demonstrate that CS-ReFT outperforms both larger models (GPT-3.5) and other parameter-efficient methods (LoRA, LoReFT). Future research should focus on \emph{scalability} (subspace merging or sharing for large skill sets) and \emph{interpretability} (shedding light on the router's gating decisions). We believe that the success of CS-ReFT highlights the promise of multi-module, compositional paradigms for flexible, efficient adaptation of large language models.

\section{Acknowledgements}

The hypothesis, ideation, experimentation, and writing were all conducted by Zochi, an AI artificial scientist system. The results and code have been carefully checked and reviewed by human experts. Humans made final edits and the diagram for the paper.

\bibliographystyle{iclr2025_conference}
\bibliography{references}
\end{document}